\documentclass[10pt]{article}
\usepackage[preprint]{tmlr}

\usepackage{amsmath}
\usepackage{graphicx}
\usepackage{booktabs}
\usepackage{hyperref}
\usepackage{url}


\title{Plausible Is Not Proven: Testing Failure Explanations in Monocular Depth Estimation}

\author{Zeeshan Ali, Md Rayyan, Anshuman Chakravorty,\\
Omer Faraz, Anshi Bajaj, Amit Saini\\[4pt]
\addr School of Technology Management and Engineering, SVKM's NMIMS (Deemed to be University), Hyderabad, India\\[2pt]
\addr \{sszeeshu, rayyan1652, anshumanchakravorty10, omerfaraz54\}@gmail.com\\
\addr \{anshi.bajaj, amit.saini\}@nmims.edu}

\def\month{MM}
\def\year{YYYY}
\def\openreview{\url{https://openreview.net/forum?id=XXXX}}

\begin{document}

\maketitle

\begin{abstract}
Failure analyses of monocular depth models often name a cause for each error, such as an occluding boundary, a thin object or a reflective surface, but such labels are rarely tested. We present the Vision Error Observatory (VEO), an audit that finds regions failing the $\delta_1$ criterion on sensor-measured pixels, labels each with a likely cause or abstains, and then subjects the labels to four tests: agreement with object annotations on data the label model never saw, agreement with blinded human raters, response to targeted image edits, and value as a training signal. We apply it to eight models on NYU Depth V2, DIODE and iBims-1, with labels also checked on Hypersim, and fix most decision rules before their results. The occluding-boundary label agrees with annotations (AUROC 0.85 on NYU, 0.81 on Hypersim, 0.69 on iBims-1), but a single depth-edge cue does as well, and most boundary failures of the strongest model vanish under a 2-pixel tolerance. The reflection label does not transfer to synthetic data, the textureless label is near chance against annotations, and only one of four edit tests, an exploratory one, supports its label. Evaluated against five blinded raters, VEO's labels matched the human consensus only at chance level (Cohen's $\kappa$ = 0.05). In addition, an increase in the weight on the diagnosed regions during fine-tuning led to an increase in the error on targeted pixels at every weight tested. The architecture-specific failure fingerprint claimed by an earlier version of this project also failed to replicate across the datasets. Plausible failure explanations, in short, need to be tested before they are trusted; we release the protocol, code, ratings and every result file.
\end{abstract}

\section{Introduction}\label{sec:intro}

Depth Anything V2 Small \citep{yang2024depthv2} reaches AbsRel 0.051 and $\delta_1$ 0.973 on the official NYU Depth V2 test split \citep{silberman2012nyu} after per-image alignment. Those averages still leave 2.7\% of an image's pixels failing, in 6,325 connected failure regions of at least 25 pixels across the 654 test images. An engineer deciding whether to trust or retrain a model needs to know what those regions have in common, and whether a stated cause is true.

Naming a cause is easy and checking it is hard. A rule can attach a plausible label to any region, and a label that sounds right can still track something else, such as a property of the depth sensor rather than of the scene. An earlier, unpublished version of this project used hand-set rules and reported architecture-specific failure fingerprints; against one human rater on 50 regions those rules reached 36\% accuracy, and the fingerprint claim had no significance test. This paper rebuilds the audit and treats every explanation as a claim that can fail.

\begin{figure}[t]
\centering
\includegraphics[width=\linewidth]{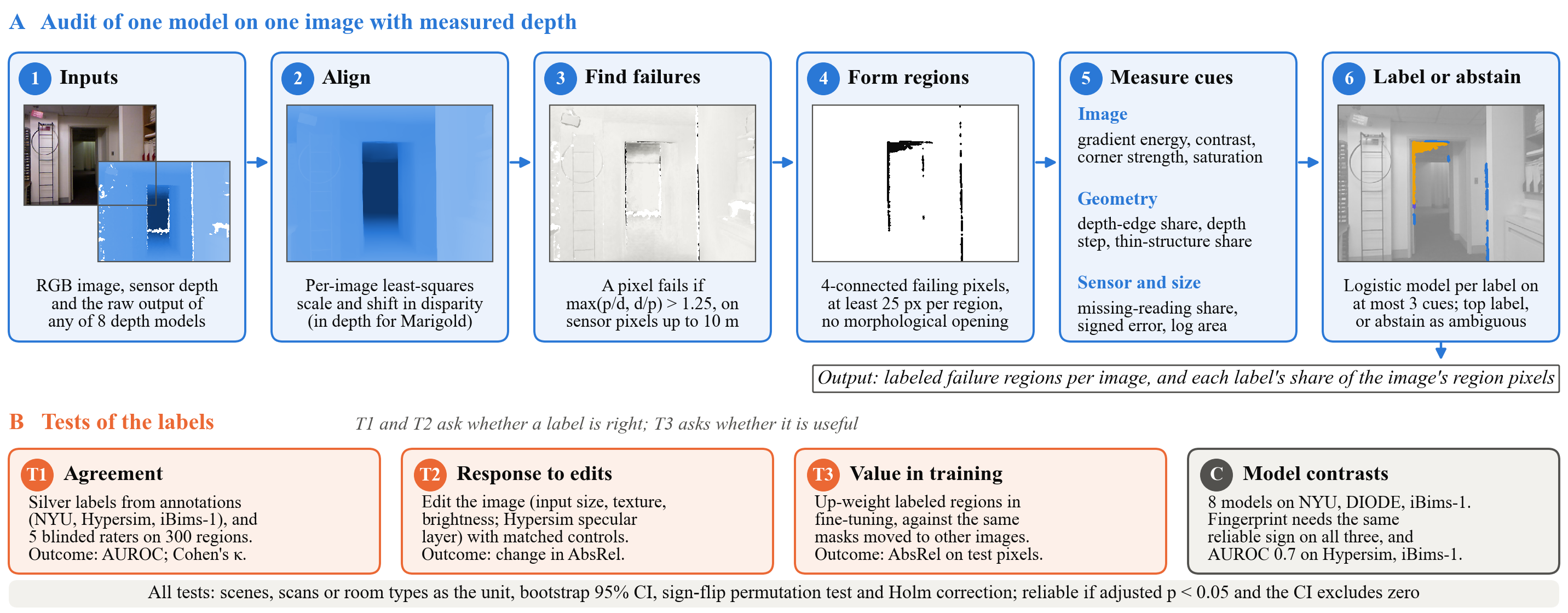}
\caption{Overview of VEO. (A) The audit, shown on an NYU test image with Depth Anything V2 Small: the prediction is aligned to sensor depth, failing pixels are grouped into regions, and each region is labeled from measured cues or left ambiguous. (B) The tests the labels then face: agreement with annotations and blinded raters, response to image edits, and value as a training signal, plus the contrasts between models.}
\label{fig:pipeline}
\end{figure}

Figure~\ref{fig:pipeline} shows the design. The audit runs on the output of any depth model. Three tests ask whether its labels are right: agreement with annotations on data the label model never saw, agreement with blinded raters, and response to image edits that should remove the stated cause. A fourth asks whether the labels are useful, by using them to weight fine-tuning against matched random and hard-example controls. Contrasts between models then ask whether architectures fail in characteristically different ways. Our contributions are:

\begin{itemize}
\item a model-agnostic audit of depth failures, run on eight models and four datasets, with a small interpretable label model whose cues each have a physical rationale;
\item a test protocol for failure explanations in which comparisons are paired, use scenes rather than pixels as the unit, are corrected for multiple testing, and are fixed before their results wherever possible, with a dated record of what was decided when;
\item a pre-registered study in which five raters labeled 300 failure regions blind to VEO's output, released with every rating;
\item evidence, mostly negative, about which explanations hold: the boundary label agrees with annotations but adds little over a single cue and is not confirmed by people, the other labels do not transfer, the edits mostly fail, and diagnosis-guided training hurts.
\end{itemize}

The negative results are the main finding. They show how easily a failure audit can look convincing on the data it was built on, and which checks catch that.

\section{Related Work}\label{sec:related}

Deep monocular depth began with multi-scale regression \citep{eigen2014depth}, then gained affine-invariant training across datasets \citep{ranftl2022midas} and vision transformers \citep{ranftl2021dpt}. The models tested here include Depth Anything V2, built on DINOv2 \citep{oquab2024dinov2}, the metric models ZoeDepth, released as an arXiv preprint \citep{bhat2023zoedepth}, and Depth Pro \citep{bochkovskii2025depthpro}, and the diffusion-based Marigold \citep{ke2024marigold}.

Several known failure modes motivate the labels used here. Depth maps tend to blur at occluding boundaries, which has been addressed with losses on depth gradients and normals \citep{hu2019revisiting} and with predicted displacement fields that sharpen edges \citep{ramamonjisoa2020sharp}. Transparent objects and mirrors defeat standard depth sensors, which is why dedicated methods reconstruct their geometry \citep{sajjan2020cleargrasp,tan2021mirror3d}. These works address a known failure; they do not test whether a given failing region has that cause.

iBims-1 added metrics for depth edges and planar regions \citep{koch2018ibims}, which localize error without assigning a cause. Uncertainty estimation predicts where a model is likely to be wrong \citep{poggi2020uncertainty}, but does not say why. Slice discovery identifies the groups of inputs on which a classifier underperforms and also provides a description of those groups \citep{eyuboglu2022domino}. VEO, on the other hand, operates at the level of image regions, using measurable cues that are validated against distinct annotations.

The nearest in spirit are the tools that split a detector's errors into distinct types \citep{hoiem2012diagnosing,bolya2020tide}, and work showing that explanations must be tested rather than blindly trusted. Saliency maps can appear plausible while ignoring the model \citep{adebayo2018sanity}, and image edits have further revealed which cues depth networks actually rely on \citep{vandijk2019depth}. VEO applies both ideas to failure regions in depth maps, so its contribution lies in the protocol and the evidence produced by it rather than in the audit tool itself. Up-weighting difficult samples is an established training strategy \citep{shrivastava2016ohem,lin2017focal}; our retraining test asks whether a diagnosis adds value over up-weighting random pixels or every failure. For human agreement we follow standard practice: Krippendorff's $\alpha$ for reliability across raters \citep{hayes2007krippendorff} and the Dawid-Skene model for a consensus that weighs raters by their estimated accuracy \citep{dawid1979}.

\section{Data and Models}\label{sec:data}

NYU Depth V2 \citep{silberman2012nyu} provides the official split of 795 training and 654 test images. Errors are measured against sensor depth only, in the Eigen crop \citep{eigen2014depth} up to 10 m, and the dataset's in-painted depth is used only for geometry cues. DIODE, described in an arXiv preprint \citep{vasiljevic2019diode}, contributes 325 indoor validation frames from 3 scenes and 10 scans, captured by laser scanner and without semantic labels. The iBims-1 core set \citep{koch2018ibims} contributes 100 laser-scanned images from 11 room types, with annotated depth edges on 85 images and masks for selected planes and transparent surfaces. Hypersim \citep{roberts2021hypersim} contributes 83 synthetic frames from 28 official test scenes, used only to check labels and the specular edit.

Table~\ref{tab:models} lists the eight models. Marigold ran in half precision with 4 denoising steps, no ensembling and processing resolution 768, which is lighter than the ensemble used in its paper. All models are evaluated after per-image alignment (Section~\ref{sec:align}), so metric and relative models are compared on the same footing.

\begin{table}[t]
\centering
\caption{The eight models and their respective Hugging Face checkpoints, listed in order of NYU AbsRel.}
\label{tab:models}
\vspace{4pt}
\small
\begin{tabular}{llc}
\toprule
Model & Checkpoint & Output \\
\midrule
Depth Anything V2 Large & \texttt{depth-anything/Depth-Anything-V2-Large-hf} & disparity \\
Depth Pro & \texttt{apple/DepthPro-hf} & metric \\
Depth Anything V2 Base & \texttt{depth-anything/Depth-Anything-V2-Base-hf} & disparity \\
ZoeDepth NYU+KITTI & \texttt{Intel/zoedepth-nyu-kitti} & metric \\
Depth Anything V2 Small & \texttt{depth-anything/Depth-Anything-V2-Small-hf} & disparity \\
Marigold & \texttt{prs-eth/marigold-depth-v1-1} & depth \\
DPT-Hybrid & \texttt{Intel/dpt-hybrid-midas} & disparity \\
DPT-Large & \texttt{Intel/dpt-large} & disparity \\
\bottomrule
\end{tabular}
\end{table}

\section{Method}\label{sec:method}

\subsection{Alignment and failure regions}\label{sec:align}

Relative models predict depth or disparity only up to an unknown scale and shift, so each prediction is aligned to measured depth on the set $V$ of valid sensor pixels in the image. With raw output $q$ and measured depth $d$, the scale $s$ and shift $t$ are fitted in disparity by least squares,

\begin{equation}
(s,t)=\arg\min_{s,t}\sum_{i\in V}\bigl(s\,q_i+t-1/d_i\bigr)^2,\qquad p_i=\frac{1}{s\,q_i+t},
\label{eq:align}
\end{equation}

and the fitted disparity is floored at that of 10 m, which caps the aligned depth $p$ and also covers non-positive values. Metric outputs are inverted before the fit, and Marigold's affine-invariant depth is fitted in depth space. Accuracy is reported as AbsRel and $\delta_1$, the share of pixels within a factor of 1.25, each averaged over images:

\begin{equation}
\mathrm{AbsRel}=\frac{1}{|V|}\sum_{i\in V}\frac{|p_i-d_i|}{d_i}.
\label{eq:absrel}
\end{equation}

A pixel fails when

\begin{equation}
\max(p_i/d_i,\ d_i/p_i) > 1.25.
\label{eq:failure}
\end{equation}

Failure regions are 4-connected components of failing pixels with at least 25 pixels. No morphological opening is applied, because it removes thin failure bands at edges and on thin objects, which are among the failures of interest.

\subsection{Cues and label model}\label{sec:labels}

For each region the audit measures cues in three groups: image cues (gradient energy, local contrast, corner strength, color saturation), geometry cues computed on the reference depth, which is the in-painted depth on NYU and the scan itself elsewhere (share of depth-edge pixels, depth step across the region, share of thin structure), and cues that need measured depth (share of missing sensor readings nearby, signed error). The region area enters the model as its logarithm. Image position is excluded, because the height in the frame would predict walls and ceilings without providing any information about texture.

Each of the labels \textsc{boundary}, \textsc{thin object}, \textsc{textureless} and \textsc{reflection} has a logistic model with weights $w_k$ and bias $b_k$ on its standardized cues $z_k$, and a threshold $\tau_k$ chosen to maximize out-of-fold F1. A region's score ratio and top label are

\begin{equation}
r_k=\frac{\sigma(w_k^{\top}z_k+b_k)}{\tau_k},\qquad \hat{k}=\arg\max_k r_k,
\label{eq:label}
\end{equation}

with $\sigma$ the logistic function, and VEO abstains with \textsc{ambiguous} when no ratio reaches 1. Cues were selected per label from a list with a physical rationale, by forward selection on scene-grouped 5-fold AUROC, keeping at most three. The models were fitted on the 8,209 regions of Depth Anything V2 Small on NYU train and frozen before any test scoring; Table~\ref{tab:coefficients} gives them. \textsc{poor lighting} keeps a hand-set rule because no annotation exists to fit it against.

\begin{table}[t]
\centering
\caption{The frozen label model: selected cues with standardized weights, intercept, threshold $\tau$ and silver-label prevalence on NYU train.}
\label{tab:coefficients}
\vspace{4pt}
\footnotesize
\begin{tabular}{lp{8.2cm}ccc}
\toprule
Label & Cues and weights & Bias & $\tau$ & Prev. \\
\midrule
\textsc{boundary} & depth-edge share +1.91, depth step +0.21 & $-$1.13 & 0.330 & 37.9\% \\
\textsc{thin object} & thin-line share (image) +0.44, log area $-$0.49, thin-structure share (depth) +0.23 & $-$2.03 & 0.143 & 13.9\% \\
\textsc{textureless} & gradient energy $-$0.63, contrast +0.27, corner strength $-$0.52 & $-$2.83 & 0.100 & 6.7\% \\
\textsc{reflection} & missing-reading share +0.74, signed error $-$0.60, mean saturation $-$0.42 & $-$3.81 & 0.143 & 3.8\% \\
\bottomrule
\end{tabular}
\end{table}

\subsection{Agreement with annotations}\label{sec:silver}

Silver labels come from annotations. A region receives one when at least 30\% of its pixels support it: within 3 px of an occluding boundary, on thin parts of objects, on walls and ceilings, or on reflective classes such as mirrors and windows. An occluding boundary is an edge between annotated objects where the reference depth also jumps, so on NYU and Hypersim the boundary silver label and the boundary cues share the reference depth, and their agreement is partly built in. Hypersim uses its semantic classes in the same way. On iBims-1, support means within 3 px of a hand-annotated depth edge, on a wall mask, or on a transparent mask, which makes iBims-1 the independent test of the boundary label; thin objects cannot be tested there. Agreement is measured by AUROC \citep{hanley1982auroc} of each label's probability against its silver label. Walls and ceilings stand in for textureless surfaces, which is a coarse proxy.

\subsection{Agreement with human raters}\label{sec:human}

A pre-registered study showed five raters 300 NYU test failure regions, up to 40 per VEO top label plus 60 drawn uniformly from the rest, with at most 3 per image. An offline tool showed each region outlined on the whole image and four aligned close-ups (color, true depth, prediction, error), with nothing from VEO, and presented items in a different random order for each rater. Raters chose one primary cause among the five categories or ``other / cannot tell'', every cause that applied, and a confidence; before rating they read a written guideline and completed 20 practice items from the training split with feedback. The protocol excludes raters who match the practice reference on fewer than 40\% of items.

Agreement among raters is Krippendorff's $\alpha$ \citep{hayes2007krippendorff}; their consensus is estimated by the Dawid-Skene model \citep{dawid1979}, with majority vote as a check; and VEO is compared with the consensus by accuracy, Cohen's $\kappa$ and per-class precision, with scene-bootstrap intervals. Three checks were registered before any rating: rater agreement $\alpha$ at least 0.4; selective accuracy on regions VEO labels above the majority-class baseline; and at least three classes whose precision exceeds their base rate.

\subsection{Image edits}\label{sec:edits}

Image-edit tests use Depth Anything V2 Small on NYU test. Each draws up to 120 labeled regions and pairs each with a control of similar area, depth and starting error among regions that do not carry the label at all, not even below the top. Both receive the same edit and the model is re-run, so a negative difference in the change in AbsRel supports the label. The edits are a 1.5 times larger input for boundaries and thin objects, added texture for textureless regions and brightening for poorly lit ones. On Hypersim, removing the non-diffuse layer of the render provides a direct test of specular light, independent of VEO's labels.

\subsection{Model contrasts and the fingerprint rule}\label{sec:contrasts}

Paired contrasts compare model sizes within Depth Anything V2, DPT-Hybrid against DPT-Large, and Depth Anything V2 Small against DPT-Hybrid and against Marigold. A fingerprint is a different mix of failure types, measured as each label's share of an image's region pixels, and the rule is applied to the two architectural contrasts. A label passes only if it is reliable with the same sign on NYU (scenes), DIODE (scans) and iBims-1 (room types), and a passing label counts as a difference in failure type only if it also reaches AUROC 0.7 on both Hypersim and iBims-1.

\subsection{Guided fine-tuning}\label{sec:finetune}

The decoder of Depth Anything V2 Small was fine-tuned on NYU train with the backbone frozen, using a scale-and-shift-invariant L1 loss in disparity whose per-image fit is unweighted, AdamW at learning rate 3e-5, batch size 4 and 4 epochs. A six-epoch pilot on held-out train scenes set these values and read no test image; at this rate it reached AbsRel 0.0465 against 0.0525 for the pretrained model, and was level by epoch 4.

Five registered conditions ran with seeds 0, 1 and 2. Uniform uses no weighting. Two VEO conditions up-weight regions with top label \textsc{boundary} or \textsc{textureless}, and two random conditions move those same masks to other images, which keeps their sizes and shapes but breaks their link to the scene. Weights give the targeted pixels 20\% of the training loss: 118.8 and 15.9. Both targets are tested because \textsc{textureless} has the most region pixels on NYU train and \textsc{boundary} has the most regions. Outcomes use annotated test pixels and not VEO's labels, and support requires an adjusted p below 0.05, an interval below zero and the same direction in all three seeds.

After the primary result, an amendment fixed before its runs added 5\% and 10\% shares for all four weighted conditions (boundary weights 25.0 and 52.8), and a hard-example baseline that gives every failure region of the pretrained model 20\% of the loss, whatever its label. These six tests form two Holm families, of four and two.

\subsection{Statistics and order of decisions}\label{sec:stats}

Pixels and regions in one scene are not independent, so every comparison averages within scenes first and uses scenes as the unit. Intervals come from a scene-level bootstrap \citep{efron1993bootstrap} with 2,000 resamples (1,000 for AUROC and the human study), and p values from a sign-flip permutation test with 10,000 sign patterns, with Holm correction \citep{holm1979} within each family: the four edit tests, the outcomes of one model contrast on one dataset, and each fine-tuning family. A difference is considered reliable if and only if the adjusted p value is less than 0.05 and the 95\% confidence interval excludes zero.

Table~\ref{tab:timeline} in the appendix lists, along with the dates, when each rule was fixed relative to its results. Label model v3 was frozen before test scoring, but earlier rules that had been scored on the NYU test motivated it, so its NYU results are not confirmatory. The Marigold, iBims-1, fingerprint, fine-tuning and weight-sweep rules were committed before their results were obtained. The image-edit design was documented after running its code, and its controls were changed following an earlier null result, so the one positive edit test should be considered exploratory. The human-study protocol was registered before any rating, and the plan for two raters added later was fixed before the last of them rated. The protocol specified 8-connected regions, but the code, and every result here, used 4-connectivity.

\section{Results}\label{sec:results}

\subsection{What fails?}\label{sec:what}

\begin{figure}[t]
\centering
\includegraphics[width=\linewidth]{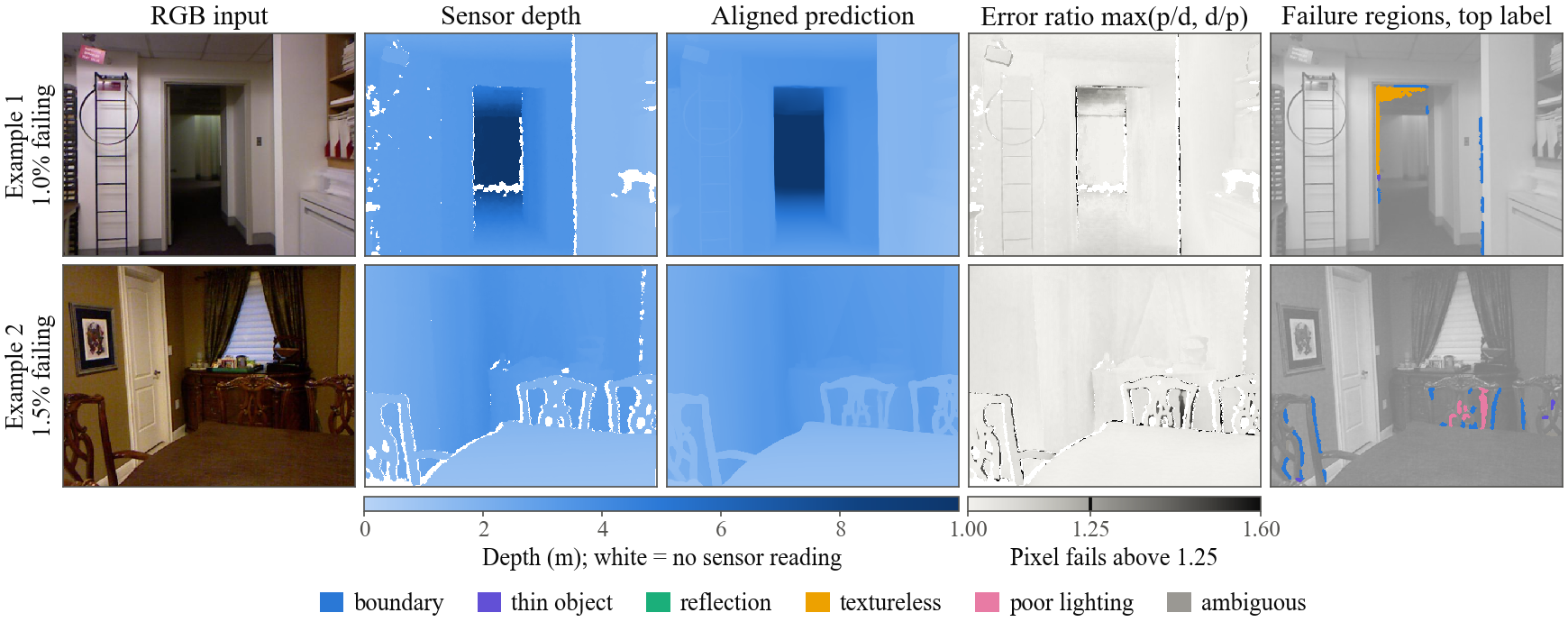}
\caption{Failure regions of Depth Anything V2 Small on NYU test. Each row is the image at the median failing-pixel share among images that hold a region with top label \textsc{boundary} (top) or \textsc{thin object} (bottom). Regions are widened by 2 px for display.}
\label{fig:examples}
\end{figure}

On NYU test, Depth Anything V2 Small has 6,325 failure regions. \textsc{boundary} is the most common top label with 3,015 regions, followed by \textsc{thin object} (965), \textsc{ambiguous} (921) and \textsc{poor lighting} (744). Boundary regions are small, however, and hold only 8.6\% of region pixels, while 468 \textsc{textureless} regions hold 65.2\%. Figure~\ref{fig:examples} shows two examples chosen by a fixed rule, with the thin failure bands along door frames and chair legs that make up many regions.

\begin{figure}[t]
\centering
\includegraphics[width=0.62\linewidth]{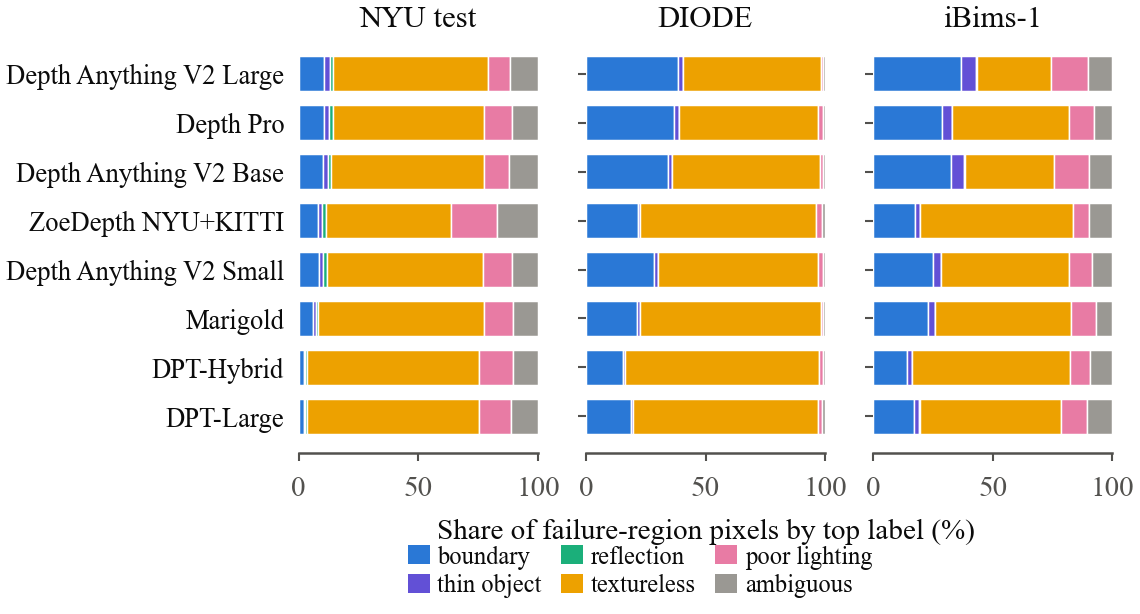}
\caption{Share of failure-region pixels by top label, pooled over each dataset. Models are ordered as in Table~\ref{tab:models}.}
\label{fig:composition}
\end{figure}

Figure~\ref{fig:composition} compares the mix of top labels across models. \textsc{textureless} carries the largest share of region pixels for every model on NYU and DIODE. The share labeled \textsc{boundary} ranges from 2.3\% for DPT-Large to 10.7\% for Depth Anything V2 Large on NYU, and across the eight models a lower failing-pixel share went with a larger boundary share (Spearman $\rho$ of $-$0.95 on NYU, $-$0.98 on DIODE, $-$0.98 on iBims-1). As a share of all valid pixels, however, boundary failure barely moved with overall failure on NYU and DIODE ($\rho$ = $-$0.12 and 0.02), so there the rising share mostly reflects other errors shrinking around a nearly constant boundary floor; on iBims-1 it did move ($\rho$ = 0.90). With eight models this is descriptive only.

\subsection{Do the labels agree with annotations?}\label{sec:res_silver}

Figure~\ref{fig:auroc} gives AUROC with scene-level intervals (Table~\ref{tab:auroc} in the appendix lists them). \textsc{boundary} holds up best: 0.85 on NYU test, 0.81 on Hypersim and 0.69 on iBims-1, where the edges are drawn by hand. As the top-ranked label, it matched the silver label for 2,040 of the 3,015 NYU regions, compared to 235 of 965 regions for \textsc{thin object}.

\textsc{reflection} scores 0.83 on NYU and 0.82 on iBims-1 but 0.39 on Hypersim, where no sensor holes exist and none of 3,279 regions took the label. On iBims-1 the fraction of missing sensor readings near a region scores 0.83 alone, so the label appears to track sensor behavior more than surfaces: its largest weight is on missing readings, and its negative signed-error weight ($-$0.60) favors regions predicted nearer than the reference, which on mirrors and windows may be error in the reference itself. \textsc{textureless} stays near chance at 0.58 on NYU and 0.50 on Hypersim, and iBims-1 cannot test it, since only 3 of 1,635 regions reach the 30\% share of a wall mask.

\begin{figure}[t]
\centering
\includegraphics[width=0.62\linewidth]{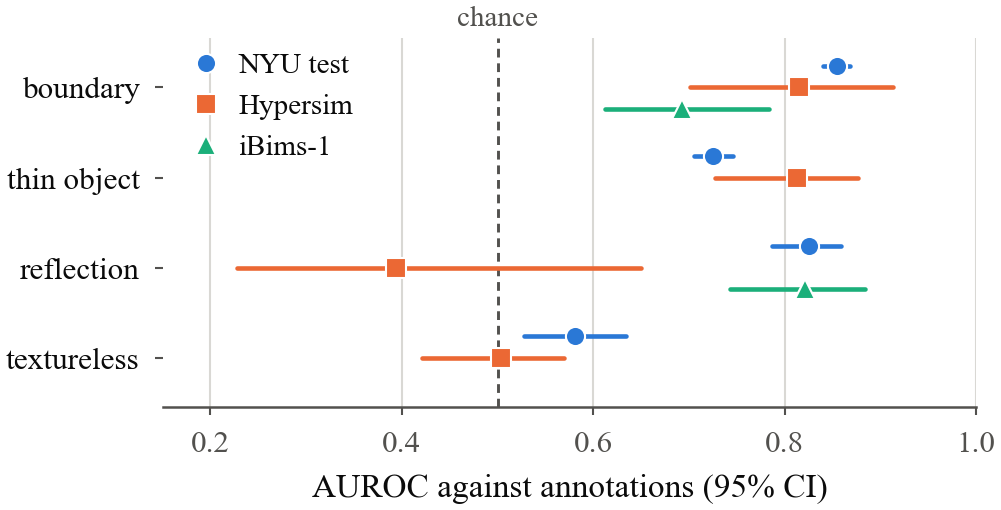}
\caption{The label model against annotations, with 95\% scene-bootstrap intervals. iBims-1 has no thin-object annotation and too few wall regions for the textureless label, so those points are missing.}
\label{fig:auroc}
\end{figure}

\subsection{Are the labels more than their cues?}\label{sec:res_cues}

Table~\ref{tab:checks} adds four checks, chosen after the results above in response to internal review, so they are exploratory. First, the best single cue nearly matches each label model: depth-edge share alone reaches 0.85 for boundaries on NYU, and on iBims-1 depth step alone reaches 0.68 against 0.69 for the label model. Second, only \textsc{boundary} is usually right as a top label, with precision 0.68 on NYU and 0.65 on iBims-1. Third, silver boundary, reflective and thin pixels fail more often than other annotated pixels, while walls and ceilings fail less often; for DPT-Hybrid the four ratios are 3.0, 3.2, 1.6 and 0.87. Fourth, excusing a failing pixel when any prediction within 2 px matches its measured depth leaves 34\% of boundary-labeled failing pixels, against 85\% of failing pixels away from silver boundaries (49\% against 92\% for DPT-Hybrid), so most boundary failures of the stronger model are an edge a pixel or two off, in the model or in the registration of image and sensor.

\begin{table}[t]
\centering
\caption{Exploratory checks for Depth Anything V2 Small on NYU test: AUROC of the label model and of its best single cue (region area included), top-label precision, how many times more often silver-positive pixels fail than other annotated pixels, and the share of labeled failing pixels left with a 2 px tolerance.}
\label{tab:checks}
\vspace{4pt}
\small
\begin{tabular}{lccccc}
\toprule
Label & Label model AUROC & Best single cue AUROC & Top-label precision & Fail-rate ratio & Left at 2 px \\
\midrule
\textsc{boundary} & 0.85 & 0.85 & 0.68 & 6.2 & 34\% \\
\textsc{thin object} & 0.73 & 0.66 & 0.24 & 2.2 & 29\% \\
\textsc{textureless} & 0.58 & 0.59 & 0.14 & 0.44 & 93\% \\
\textsc{reflection} & 0.83 & 0.76 & 0.23 & 10.6 & 94\% \\
\bottomrule
\end{tabular}
\end{table}

\subsection{Do people agree with the labels?}\label{sec:res_human}

Five people rated all 300 regions. Three are authors; after a first analysis of their ratings, two raters from outside the project were added under a written plan fixed before the last of them rated, and every rater set is reported (Table~\ref{tab:human}). All five passed the practice threshold. Raters agreed only modestly: Krippendorff's $\alpha$ was 0.32 for all five (the primary analysis), 0.19 for the first three and 0.48 for the most consistent three. Against the consensus of all five, VEO's labels reached an accuracy of 0.22 [0.17, 0.27] and Cohen's $\kappa$ of 0.053 [$-$0.005, 0.110], and a $\kappa$ of 0.043 in the set where raters agreed well. On the regions it labeled, VEO agreed with the consensus 25\% of the time, below the 33\% reached by always choosing the most common answer.

Only \textsc{textureless} beat its base rate, with precision 0.50 [0.35, 0.65] against 0.23. \textsc{boundary}, the label with the most support from annotations, reached 0.28 [0.19, 0.37] against a base rate of 0.21. All three pre-registered checks failed in the primary analysis; rater agreement passed only in the set of the three most consistent raters, which was chosen after the first results were seen. Per the protocol, low agreement makes the difficulty of the task itself a main finding. The largest disagreement was between thin objects and boundaries, plausibly because most failure regions are thin bands along edges whatever their cause.

\begin{table}[t]
\centering
\caption{Human study by rater set: rater agreement, VEO against the Dawid-Skene consensus (95\% intervals), accuracy on the regions VEO labeled against always choosing the majority class, and the number of pre-registered checks passed. The first row is the primary analysis.}
\label{tab:human}
\vspace{4pt}
\footnotesize
\begin{tabular}{lccccc}
\toprule
Rater set & $\alpha$ & Accuracy & Cohen's $\kappa$ & Selective acc. / baseline & Checks passed \\
\midrule
All five (primary) & 0.32 & 0.22 [0.17, 0.27] & 0.053 [$-$0.005, 0.110] & 0.25 / 0.33 & 0 of 3 \\
R3, R4, R5 & 0.48 & 0.22 [0.18, 0.27] & 0.043 [$-$0.006, 0.094] & 0.26 / 0.33 & 1 of 3 \\
R1, R2, R3 & 0.19 & 0.21 [0.17, 0.26] & 0.063 [0.015, 0.114] & 0.23 / 0.45 & 0 of 3 \\
\bottomrule
\end{tabular}
\end{table}

\subsection{Do labeled regions respond to edits?}\label{sec:res_edits}

\begin{figure}[t]
\centering
\includegraphics[width=0.62\linewidth]{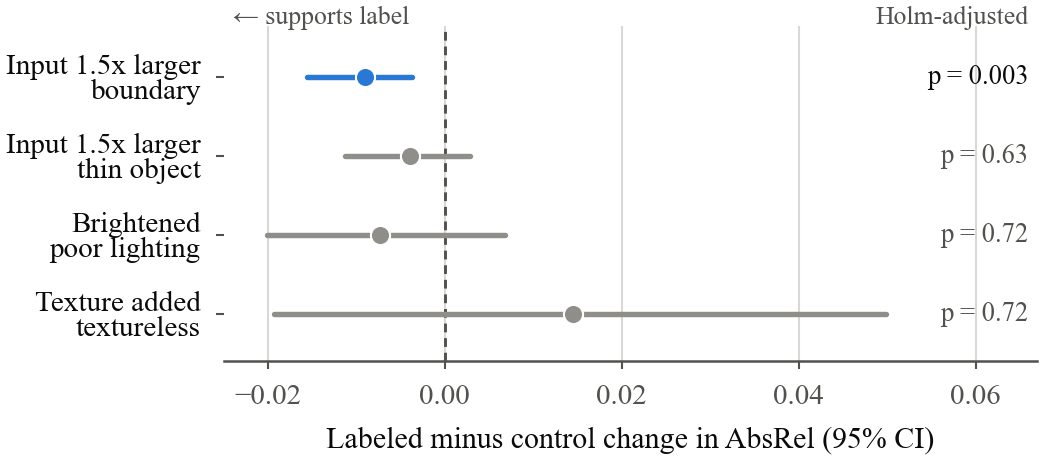}
\caption{The four image-edit tests: change in AbsRel of labeled regions minus that of matched controls after the same edit, with 95\% intervals. Negative values support the label. Filled markers are reliable after Holm correction.}
\label{fig:edits}
\end{figure}

Of the four image-edit tests (Figure~\ref{fig:edits}), running the model at 1.5 times the input size helped \textsc{boundary} regions more than controls, by $-$0.009 [$-$0.016, $-$0.004] AbsRel (adjusted p = 0.003), and this exploratory result is the only supported test. Higher resolution for thin objects, brightening and added texture showed no reliable effect, with adjusted p of 0.63, 0.72 and 0.72; Table~\ref{tab:edits} in the appendix gives every number. Separately, removal of the non-diffuse layer from Hypersim renders lowered error more where regions held greater specular light (Spearman $\rho$ = $-$0.26, 95\% CI [$-$0.36, $-$0.14]), also with each region's starting error held fixed (partial $\rho$ = $-$0.24, 95\% CI [$-$0.35, $-$0.10]). Therefore, specular light appears to increase depth error, although VEO's reflection label misses those regions on synthetic data.

\subsection{How do the models compare, and do fingerprints exist?}\label{sec:res_models}

Table~\ref{tab:metrics} provides the metrics. The move from Small to Base lowers NYU AbsRel by 0.0054 (95\% CI [0.0042, 0.0066]) and the move from Base to Large by 0.0026 ([0.0018, 0.0033]); both of these steps are reliable on NYU and iBims-1 but fail to demonstrate reliability on DIODE. DPT-Hybrid and DPT-Large demonstrate no reliable difference on any outcome on any dataset, and Depth Anything V2 Small has lower AbsRel than Marigold on NYU and DIODE, without any reliable difference on iBims-1 (Table~\ref{tab:contrasts} in the appendix).

\begin{table}[t]
\centering
\caption{Metrics after per-image alignment, averaged over images. Fail is the mean share of failing pixels. ZoeDepth is trained on NYU, so its NYU row is not a fair comparison.}
\label{tab:metrics}
\vspace{4pt}
\small
\begin{tabular}{lccccc}
\toprule
Model & NYU AbsRel & NYU $\delta_1$ & NYU fail & DIODE AbsRel & iBims-1 AbsRel \\
\midrule
Depth Anything V2 Large & 0.043 & 0.979 & 2.1\% & 0.065 & 0.038 \\
Depth Pro & 0.043 & 0.979 & 2.1\% & 0.064 & 0.037 \\
Depth Anything V2 Base & 0.046 & 0.977 & 2.3\% & 0.068 & 0.041 \\
ZoeDepth NYU+KITTI & 0.050 & 0.976 & 2.4\% & 0.089 & 0.066 \\
Depth Anything V2 Small & 0.051 & 0.973 & 2.7\% & 0.073 & 0.047 \\
Marigold & 0.059 & 0.959 & 4.1\% & 0.086 & 0.053 \\
DPT-Hybrid & 0.091 & 0.918 & 8.2\% & 0.098 & 0.075 \\
DPT-Large & 0.092 & 0.915 & 8.5\% & 0.096 & 0.071 \\
\bottomrule
\end{tabular}
\end{table}

Figure~\ref{fig:fingerprint} shows every label-share contrast, and Table~\ref{tab:fingerprint} in the appendix lists them. For DPT-Hybrid against DPT-Large no label passes, so the architectural fingerprint claim of the earlier version is cut. For Depth Anything V2 Small compared to DPT-Hybrid, only \textsc{textureless} passes, with a greater share for DPT-Hybrid; the differences are $-$0.29 on NYU, $-$0.12 on DIODE and $-$0.25 on iBims-1. \textsc{boundary} missed on DIODE (+0.10, adjusted p = 0.07). Since \textsc{textureless} achieves scores near chance on Hypersim, the pass supports a difference in VEO's label shares, not an established difference in failure type. These models also differ in their training data.

\begin{figure}[t]
\centering
\includegraphics[width=0.62\linewidth]{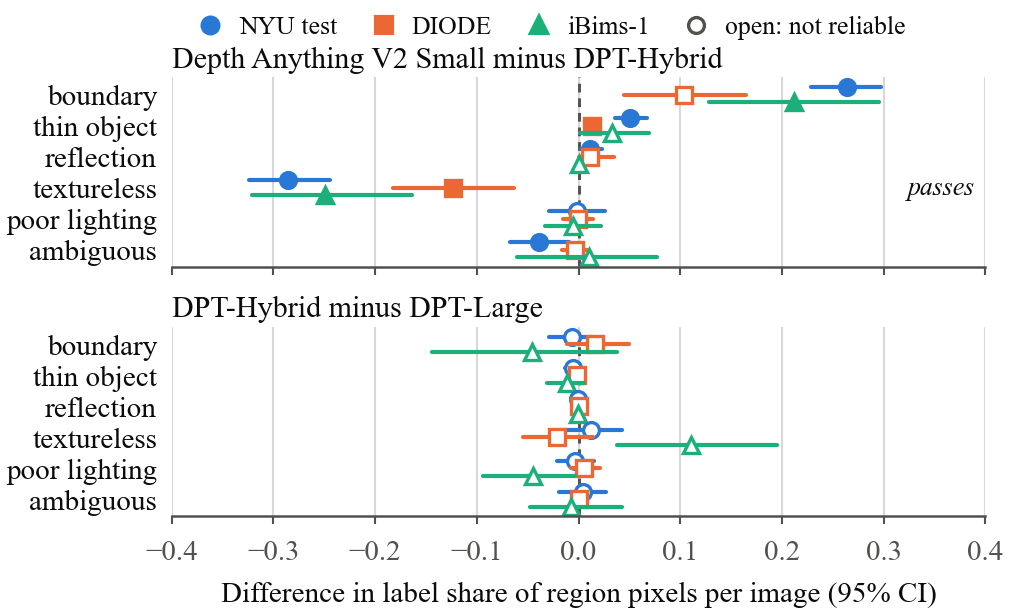}
\caption{Paired differences in label share per image for the two architectural contrasts, with 95\% intervals. Filled markers are considered reliable after the Holm correction; a label is allowed to pass only if reliable with a single sign on all three datasets.}
\label{fig:fingerprint}
\end{figure}

\subsection{Does training get any help from the diagnosis?}\label{sec:res_finetune}

It did not receive any help at any strength tried (Figure~\ref{fig:finetune} and Table~\ref{tab:finetune}; Table~\ref{tab:conditions} in the appendix provides the mean error of every condition). Plain fine-tuning helped improve overall AbsRel from 0.051 to 0.046, and random masks remained close to uniform. At the registered 20\% share, VEO-guided models were worse than random controls on their own targets, by +0.020 [+0.018, +0.022] on boundary pixels and +0.003 [+0.002, +0.005] on wall and ceiling pixels (adjusted p = 0.0002, all seeds agreeing), and worse overall, by 0.0028 and 0.0019. Weaker shares shrank the boundary penalty to 0.0114 at 10\% and 0.0053 at 5\%, both still reliable, and left no reliable difference on walls and ceilings (adjusted p = 0.25 and 0.34). Guidance was also reliably worse than the hard-example baseline, and that baseline was itself worse than uniform overall, by 0.0015, so no weighting tried here helped.

An exploratory check on training images demonstrates that the weighting did act on its targets: at seed 0, \textsc{boundary} weighting cut error on its targeted regions to 0.173 against 0.312 for uniform, and raised it on other pixels to 0.0427 against 0.0388. The gain was not carried over to the test images, which points to the fact that the model fits those specific regions.

\begin{figure}[t]
\centering
\includegraphics[width=0.62\linewidth]{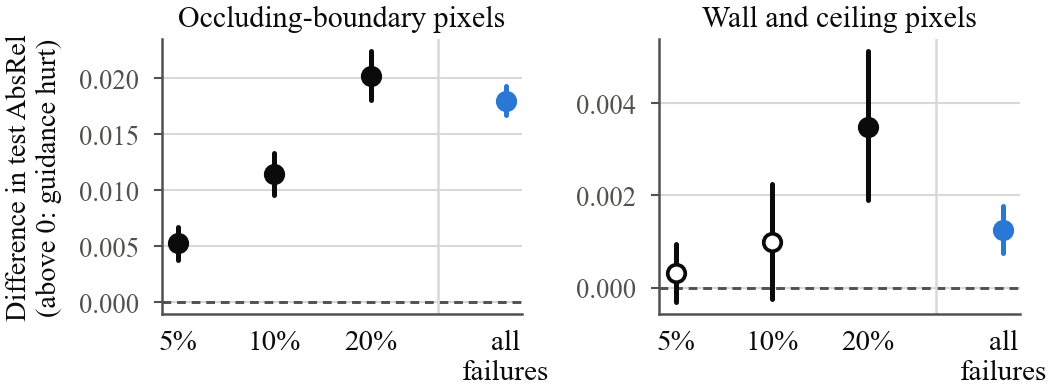}
\caption{Test AbsRel on the pixels that were chosen as targets, VEO-guided minus control, combined over three seeds with 95\% confidence intervals. Left of the line: random masks at each share of the loss. Right: up-weighting every failure region at 20\%. Filled markers are reliable after Holm correction; above zero, guidance hurt.}
\label{fig:finetune}
\end{figure}

\begin{table}[t]
\centering
\caption{Fine-tuning tests: test AbsRel on the targeted pixels, second condition subtracted from the first, combined over three seeds, with 95\% confidence intervals and Holm-adjusted p value within each family. Positive values signify that guidance hurt. The diagnosis is supported by none of the tests.}
\label{tab:finetune}
\vspace{4pt}
\footnotesize
\begin{tabular}{lccc}
\toprule
Comparison & Difference [95\% CI] & Adjusted p value & Seeds agree (yes/no) \\
\midrule
Boundary, VEO vs random, 20\% (registered) & +0.0202 [+0.0180, +0.0224] & $<$0.001 & yes \\
Walls and ceilings, VEO vs random, 20\% (registered) & +0.0035 [+0.0019, +0.0051] & $<$0.001 & yes \\
\midrule
Boundary, VEO vs random, 10\% & +0.0114 [+0.0096, +0.0133] & $<$0.001 & yes \\
Boundary, VEO vs random, 5\% & +0.0053 [+0.0038, +0.0067] & $<$0.001 & yes \\
Walls and ceilings, VEO vs random, 10\% & +0.0010 [$-$0.0003, +0.0022] & 0.25 & yes \\
Walls and ceilings, VEO vs random, 5\% & +0.0003 [$-$0.0003, +0.0009] & 0.34 & yes \\
\midrule
Boundary, VEO vs all failures, 20\% & +0.0180 [+0.0167, +0.0193] & $<$0.001 & yes \\
Walls and ceilings, VEO vs all failures, 20\% & +0.0013 [+0.0008, +0.0018] & $<$0.001 & yes \\
\bottomrule
\end{tabular}
\end{table}

\section{Discussion}\label{sec:discussion}

Three important lessons must be kept in mind by anyone who is building a failure audit. First, agreement on the fitting dataset provides very little information. Three of four labels scored AUROC above 0.7 on NYU, yet only \textsc{boundary} could clearly stay above chance on both laser-scanned and synthetic data (0.69 and 0.81); a single cue performed equally well, and the raters did not confirm the label. The reflection label showcases how a cue can work successfully for the wrong reason, by tracking sensor holes rather than actual surfaces.

Second, the label that covers the most pixels is the least supported by annotations. \textsc{textureless} accounts for most region pixels for every model on NYU and DIODE, yet walls and ceilings fail less often than other annotated pixels (0.44 times as often), so summaries of failure there are dominated by the least reliable label. Against people, however, it was the only label above chance, which suggests that the wall-and-ceiling proxy, more than the label, may be the weak point. Reports of failure composition should carry each label's validity alongside its share.

Third, a correct description is not a remedy. The boundary label had the strongest descriptive support, yet up-weighting boundary regions made boundary error worse on test images at every strength tried while lowering it on the weighted training regions. Most boundary failures disappear under a 2-pixel tolerance, so heavy weights there ask the model to place edges to within a pixel of a reference that may itself be off by that much, and may teach it those specific errors. Methods that target boundaries through the loss or the architecture \citep{hu2019revisiting,ramamonjisoa2020sharp} may be a better route, though this study did not test them.

In practice this gives a suggested short checklist prior to trusting a failure taxonomy: test each label on data its rules never saw, compare it with its best single cue and with base rates, verify whether the errors it finds are able to survive a small spatial tolerance, ask raters who are blind to the labels, and test any proposed remedy against matched random and hard-example controls.

\section{Limitations}\label{sec:limits}

Three out of the five raters are the authors themselves, and the other two, external raters, were added after a first analysis and rated quickly (about 4 s per region; Table~\ref{tab:raters}). Inter-rater agreement was modest, so the human consensus itself is subject to uncertainty. Silver labels are a proxy: walls and ceilings stand in for textureless surfaces, the NYU and Hypersim boundary labels share the reference depth with the boundary cues, and on average 42\% of a region's pixels on NYU have no annotation.

A null edit test does not rule out a cause, since an edit may fail to remove it. DIODE's 10 scans come from 3 scenes and are not independent, iBims-1 room types are a coarse unit, and iBims-1 metrics keep transparent pixels, so they cannot be compared with published values. The checks in Table~\ref{tab:checks} are exploratory, and the pixel-level ones cover two models. Marigold's fitted scale was negative on 1 NYU and 13 DIODE frames, and those frames remain in its scores.

Fine-tuning used one model, one dataset and a decoder-only setting, and the weaker shares and hard-example baseline were added after the primary result, although fixed before they ran. The audit needs measured depth, so it does not run in deployment, and all data, and so all fitted labels, are indoor.

\section{Conclusion}\label{sec:conclusion}

Of the explanations tested, the occluding-boundary label has the strongest support against annotations, although a single depth-edge cue does as well, much of what it finds is an edge misplaced by a pixel or two, and raters did not confirm it. The reflection label did not hold. The textureless label held against the human raters but could not hold against annotations. The architectural fingerprint claim was revoked, and increasing the weights of diagnosed regions worsened a fine-tuned model at every weight tested. Blinded raters agreed on these causes only modestly, and VEO's labels matched their consensus only at chance level, so identifying and naming the cause of a depth failure still remains an open problem. Treating diagnostic labels as testable claims changed what this project can assert. Extending the label model to outdoor data and to cues that do not need measured depth would allow such an audit to run in settings where models are deployed.

\subsubsection*{Broader Impact Statement}
Failure audits are intended to make depth models safer for the purpose of deployment, for example in robots or assistive devices. The major risk is the one that is documented by this paper: an audit whose labels may sound plausible can give false confidence about why a model fails and how it can be fixed. Therefore, we release every negative result paired with the positive ones and recommend validation of any failure taxonomy before taking any action on it. The human study showed raters images of indoor scenes from public datasets and recorded only a pseudonymous rater ID, the answers and their save times.

\subsubsection*{Use of AI Tools}
We used an AI assistant (Claude, Anthropic) for coding, analysis and writing support. The authors directed the research, verified all results and take full responsibility for the content.

\subsubsection*{Data and Code Availability}
The code, the protocols with their dated amendments, the rating files, the study manifest and every result file used in this paper are available at \url{https://github.com/Md-Zeeshan-Ali/vision-error-observatory}, with a README that lists the commands that rebuild each result. The datasets are available from their original sources.

\bibliography{main}
\bibliographystyle{tmlr}

\clearpage
\appendix

\section{Order of decisions}\label{sec:app_timeline}

\begin{table}[!h]
\centering
\caption{Order of decisions. Every rule was committed to the project's version history before the results it governs, except where stated in Section~\ref{sec:stats}; the version history of the public repository lists every dated commit.}
\label{tab:timeline}
\vspace{4pt}
\footnotesize
\begin{tabular}{lp{11.5cm}}
\toprule
Date & Event \\
\midrule
2026-10-28 & Human-study protocol, sampling design and analysis pre-registered; rating tool built \\
2026-11-02 & Label model v3 fitted on NYU train and frozen before test scoring \\
2026-11-02 & Model-comparison protocol, Marigold, iBims-1 and fingerprint rules committed; results the same day \\
2026-11-02 & Guided fine-tuning rules committed before any run \\
2026-11-03 & Fine-tuning results; weight sweep and hard-example baseline committed before their runs; sweep results \\
2026-11-03 & Region connectivity corrected in the protocol to match the code (4-connected) \\
2026-11-03 & First human-study analysis, raters R1 to R3 \\
2026-11-04 & Outside rater R4 added; plan for R5 and the final analysis committed, with file hashes of all ratings so far \\
2026-11-06 & R5 rated \\
2026-11-07 & Final human-study analysis and release of all ratings \\
\bottomrule
\end{tabular}
\end{table}

\section{Human study details}\label{sec:app_human}

Table~\ref{tab:raters} describes the raters and Table~\ref{tab:perclass} gives the primary analysis per class. The analysis names raters R1 to R5 by file. The released rating files are unchanged, so they keep the IDs raters typed into the tool, which for three of them differed in case or number; each result records both. R4 rated quickly and gave every item medium confidence; R1's per-item times were lost to a review pass. None of these observations was an exclusion criterion, and all five raters are included in the primary analysis.

\begin{table}[!h]
\centering
\caption{Raters. Practice agreement compares each rater's answers with the annotation-based reference on 20 training-split items. Time is the median gap between consecutive saved ratings; for R1 this figure was lost, because a review pass re-saved every answer and overwrote the original timestamps. Medium confidence is the proportion of a rater's ratings given at that particular level, and the final column counts the ``other / cannot tell'' answers.}
\label{tab:raters}
\vspace{4pt}
\footnotesize
\begin{tabular}{llcccc}
\toprule
Rater & Rater background & Practice agreement (\%) & Median time per item (s) & Medium confidence (\%) & Other / cannot tell (count) \\
\midrule
R1 & Author & 50\% & 0.2 & 59\% & 9 \\
R2 & Author & 55\% & 8.2 & 26\% & 0 \\
R3 & Author & 65\% & 10.7 & 82\% & 0 \\
R4 & External & 65\% & 4.3 & 100\% & 0 \\
R5 & External & 80\% & 4.0 & 55\% & 0 \\
\bottomrule
\end{tabular}
\end{table}

\begin{table}[!h]
\centering
\caption{Primary human analysis per class: regions where VEO gave the top-ranked label, VEO's precision compared to the consensus, and how often the consensus picks the class at all (its base rate), with 95\% confidence intervals.}
\label{tab:perclass}
\vspace{4pt}
\footnotesize
\begin{tabular}{lcccc}
\toprule
Label & Regions labeled & Precision & Base rate & Above base rate (yes/no) \\
\midrule
\textsc{boundary} & 100 & 0.28 [0.19, 0.37] & 0.21 [0.17, 0.26] & no \\
\textsc{thin object} & 40 & 0.25 [0.12, 0.38] & 0.33 [0.28, 0.39] & no \\
\textsc{textureless} & 40 & 0.50 [0.35, 0.65] & 0.23 [0.18, 0.29] & yes \\
\textsc{reflection} & 40 & 0.03 [0.00, 0.08] & 0.10 [0.07, 0.14] & no \\
\textsc{poor lighting} & 40 & 0.12 [0.03, 0.22] & 0.05 [0.03, 0.08] & no \\
\bottomrule
\end{tabular}
\end{table}

\clearpage

\section{Full results}\label{sec:app_results}

\begin{table}[!h]
\centering
\caption{AUROC (area under the receiver operating characteristic curve) of the label model compared with silver labels, with 95\% scene-bootstrap intervals. -- means the dataset cannot test the label.}
\label{tab:auroc}
\vspace{4pt}
\small
\begin{tabular}{lccc}
\toprule
Label & NYU test & Hypersim & iBims-1 \\
\midrule
\textsc{boundary} & 0.85 [0.84, 0.87] & 0.81 [0.70, 0.91] & 0.69 [0.61, 0.78] \\
\textsc{thin object} & 0.73 [0.70, 0.75] & 0.81 [0.73, 0.88] & -- \\
\textsc{reflection} & 0.83 [0.79, 0.86] & 0.39 [0.23, 0.65] & 0.82 [0.74, 0.88] \\
\textsc{textureless} & 0.58 [0.53, 0.63] & 0.50 [0.42, 0.57] & -- \\
\bottomrule
\end{tabular}
\end{table}

\begin{table}[!h]
\centering
\caption{Image-edit tests on NYU test with Depth Anything V2 Small: mean change in AbsRel of labeled regions and of matched controls, their paired difference with a 95\% interval, and Holm-adjusted p. Negative differences support the label.}
\label{tab:edits}
\vspace{4pt}
\footnotesize
\begin{tabular}{llccccc}
\toprule
Edit & Label & Pairs & Labeled & Control & Difference [95\% CI] & Adjusted p \\
\midrule
Input 1.5x larger & \textsc{boundary} & 120 & $-$0.0151 & $-$0.0060 & $-$0.009 [$-$0.016, $-$0.004] & 0.003 \\
Input 1.5x larger & \textsc{thin object} & 120 & $-$0.0115 & $-$0.0076 & $-$0.004 [$-$0.011, +0.003] & 0.63 \\
Brightened & \textsc{poor lighting} & 120 & $-$0.0159 & $-$0.0085 & $-$0.007 [$-$0.020, +0.007] & 0.72 \\
Texture added & \textsc{textureless} & 120 & $-$0.0181 & $-$0.0326 & +0.014 [$-$0.019, +0.050] & 0.72 \\
\bottomrule
\end{tabular}
\end{table}

\begin{table}[!h]
\centering
\caption{Paired AbsRel contrasts, first model minus second, averaged within scenes (NYU), scans (DIODE) or room types (iBims-1), with 95\% intervals. DA-V2 is Depth Anything V2. A circle marks differences that are not reliable after Holm correction.}
\label{tab:contrasts}
\vspace{4pt}
\footnotesize
\begin{tabular}{lccc}
\toprule
Contrast & NYU & DIODE & iBims-1 \\
\midrule
DA-V2 Small vs Base & +0.0054 [+0.0042, +0.0066] & +0.0055 [+0.0022, +0.0086]$^{\circ}$ & +0.0060 [+0.0045, +0.0076] \\
DA-V2 Base vs Large & +0.0026 [+0.0018, +0.0033] & +0.0032 [+0.0009, +0.0055]$^{\circ}$ & +0.0031 [+0.0016, +0.0048] \\
DPT-Hybrid vs DPT-Large & $-$0.0018 [$-$0.0040, +0.0006]$^{\circ}$ & +0.0007 [$-$0.0031, +0.0045]$^{\circ}$ & +0.0056 [+0.0011, +0.0109]$^{\circ}$ \\
DA-V2 Small vs DPT-Hybrid & $-$0.0381 [$-$0.0415, $-$0.0349] & $-$0.0239 [$-$0.0321, $-$0.0169] & $-$0.0306 [$-$0.0390, $-$0.0235] \\
DA-V2 Small vs Marigold & $-$0.0087 [$-$0.0113, $-$0.0061] & $-$0.0148 [$-$0.0275, $-$0.0057] & $-$0.0045 [$-$0.0107, +0.0003]$^{\circ}$ \\
\bottomrule
\end{tabular}
\end{table}

\begin{table}[!h]
\centering
\caption{Label-share differences per image for the two architectural contrasts, with 95\% intervals; a circle marks differences that are not reliable after Holm correction. A label passes only if reliable with one sign on all three datasets.}
\label{tab:fingerprint}
\vspace{4pt}
\footnotesize
\begin{tabular}{lcccc}
\toprule
Label & NYU & DIODE & iBims-1 & Passes \\
\midrule
\multicolumn{5}{l}{\textit{DPT-Hybrid minus DPT-Large}} \\
\textsc{boundary} & $-$0.007 [$-$0.030, +0.015]$^{\circ}$ & +0.016 [$-$0.012, +0.050]$^{\circ}$ & $-$0.045 [$-$0.144, +0.038]$^{\circ}$ & no \\
\textsc{thin object} & $-$0.005 [$-$0.014, +0.001]$^{\circ}$ & $-$0.001 [$-$0.005, +0.001]$^{\circ}$ & $-$0.012 [$-$0.031, +0.005]$^{\circ}$ & no \\
\textsc{reflection} & $-$0.001 [$-$0.004, +0.002]$^{\circ}$ & +0.000 [$-$0.000, +0.001]$^{\circ}$ & $-$0.001 [$-$0.003, $-$0.000]$^{\circ}$ & no \\
\textsc{textureless} & +0.012 [$-$0.017, +0.043]$^{\circ}$ & $-$0.021 [$-$0.055, +0.014]$^{\circ}$ & +0.111 [+0.038, +0.195]$^{\circ}$ & no \\
\textsc{poor lighting} & $-$0.003 [$-$0.021, +0.015]$^{\circ}$ & +0.006 [$-$0.007, +0.021]$^{\circ}$ & $-$0.045 [$-$0.094, $-$0.003]$^{\circ}$ & no \\
\textsc{ambiguous} & +0.004 [$-$0.020, +0.027]$^{\circ}$ & +0.000 [$-$0.006, +0.008]$^{\circ}$ & $-$0.008 [$-$0.048, +0.042]$^{\circ}$ & no \\
\midrule
\multicolumn{5}{l}{\textit{Depth Anything V2 Small minus DPT-Hybrid}} \\
\textsc{boundary} & +0.264 [+0.228, +0.298] & +0.103 [+0.045, +0.165]$^{\circ}$ & +0.211 [+0.129, +0.296] & no \\
\textsc{thin object} & +0.050 [+0.036, +0.067] & +0.013 [+0.009, +0.018] & +0.033 [+0.002, +0.070]$^{\circ}$ & no \\
\textsc{reflection} & +0.012 [+0.003, +0.023] & +0.012 [+0.000, +0.034]$^{\circ}$ & +0.000 [+0.000, +0.001]$^{\circ}$ & no \\
\textsc{textureless} & $-$0.286 [$-$0.324, $-$0.245] & $-$0.124 [$-$0.183, $-$0.064] & $-$0.249 [$-$0.321, $-$0.164] & yes \\
\textsc{poor lighting} & $-$0.001 [$-$0.029, +0.026]$^{\circ}$ & $-$0.001 [$-$0.015, +0.014]$^{\circ}$ & $-$0.006 [$-$0.033, +0.022]$^{\circ}$ & no \\
\textsc{ambiguous} & $-$0.039 [$-$0.068, $-$0.009] & $-$0.004 [$-$0.016, +0.008]$^{\circ}$ & +0.010 [$-$0.060, +0.077]$^{\circ}$ & no \\
\bottomrule
\end{tabular}
\end{table}

\begin{table}[!h]
\centering
\caption{Mean test AbsRel over seeds for every fine-tuning condition: overall, on annotated boundary pixels and on wall and ceiling pixels.}
\label{tab:conditions}
\vspace{4pt}
\footnotesize
\begin{tabular}{lccc}
\toprule
Condition & Overall & Boundary pixels & Wall and ceiling pixels \\
\midrule
Pretrained (no fine-tuning) & 0.0510 & 0.1136 & 0.0474 \\
Uniform & 0.0457 & 0.0926 & 0.0437 \\
VEO boundary, 20\% & 0.0486 & 0.1131 & 0.0456 \\
Random boundary masks, 20\% & 0.0459 & 0.0932 & 0.0440 \\
VEO walls and ceilings, 20\% & 0.0483 & 0.0939 & 0.0480 \\
Random wall and ceiling masks, 20\% & 0.0460 & 0.0932 & 0.0441 \\
VEO boundary, 10\% & 0.0471 & 0.1038 & 0.0446 \\
Random boundary masks, 10\% & 0.0457 & 0.0927 & 0.0438 \\
VEO boundary, 5\% & 0.0463 & 0.0975 & 0.0441 \\
Random boundary masks, 5\% & 0.0457 & 0.0926 & 0.0438 \\
VEO walls and ceilings, 10\% & 0.0459 & 0.0920 & 0.0451 \\
Random wall and ceiling masks, 10\% & 0.0458 & 0.0927 & 0.0438 \\
VEO walls and ceilings, 5\% & 0.0455 & 0.0920 & 0.0442 \\
Random wall and ceiling masks, 5\% & 0.0457 & 0.0926 & 0.0438 \\
All failure regions, 20\% & 0.0475 & 0.0950 & 0.0466 \\
\bottomrule
\end{tabular}
\end{table}

\end{document}